\documentclass[conference]{IEEEtran}
\IEEEoverridecommandlockouts
\usepackage{cite}
\usepackage{amsmath,amssymb,amsfonts}
\usepackage{graphicx}
\usepackage{textcomp}
\usepackage{xcolor}
\usepackage{algorithm,algorithmicx,algpseudocode}
\usepackage{graphics}
\usepackage{adjustbox}
\usepackage{amsfonts}
\usepackage{chngcntr}
\usepackage{booktabs} 
\usepackage{multirow} 
\usepackage{booktabs} 
\usepackage{pifont}   
\usepackage[colorlinks=true, urlcolor=red]{hyperref}
\usepackage{enumitem}
\usepackage{makecell}
\usepackage[misc]{ifsym}

\makeatletter
\algnewcommand{\LineComment}[1]{\Statex \hskip\ALG@thistlm \(\triangleright\) #1}
\makeatother

\def\BibTeX{{\rm B\kern-.05em{\sc i\kern-.025em b}\kern-.08em
    T\kern-.1667em\lower.7ex\hbox{E}\kern-.125emX}}
\begin{document}

\title{Reinforcement Learning-based Token Pruning in Vision Transformers: A Markov Game Approach}

\author{\IEEEauthorblockN{Chenglong Lu\IEEEauthorrefmark{1}, Shen Liang\IEEEauthorrefmark{2}\Letter, Xuewei Wang\IEEEauthorrefmark{3}, Wei Wang\IEEEauthorrefmark{1}}
	
	\IEEEauthorblockA{\IEEEauthorrefmark{1}School of Computer Science, Fudan University, China\\}
	\IEEEauthorblockA{\IEEEauthorrefmark{2}Data Intelligence Institute of Paris (diiP) \& LIPADE, Université Paris Cité, Paris, France\\ 
	}
	\IEEEauthorblockA{\IEEEauthorrefmark{3}State Key Laboratory of Mechanical Behavior and System Safety of Traffic Engineering,\\ Shijiazhuang Tiedao University, Shijiazhuang, China\\}
	cllu19@fudan.edu.cn, shen.liang@u-paris.fr, xwwang@stdu.edu.cn, weiwang1@fudan.edu.cn
}
\vspace{-0.4cm}
\maketitle

\maketitle

\begin{abstract}

    Vision Transformers (ViTs) have computational costs scaling quadratically with the number of tokens, calling for effective token pruning policies. Most existing policies are handcrafted, lacking adaptivity to varying inputs. Moreover, they fail to consider the sequential nature of token pruning across multiple layers. In this work, for the first time (as far as we know), we exploit \textit{Reinforcement Learning (RL)} to data-adaptively \textit{learn} a pruning policy. Formulating token pruning as a sequential decision-making problem, we model it as a \textit{Markov Game} and utilize \textit{Multi-Agent Proximal Policy Optimization (MAPPO)} where each agent makes an individualized pruning decision for a single token. We also develop reward functions that enable simultaneous collaboration and competition of these agents to balance efficiency and accuracy. On the well-known ImageNet-1k dataset, our method improves the inference speed by up to 44\% while incurring only a negligible accuracy drop of 0.4\%. The source code is available at \url{https://github.com/daashuai/rl4evit}.

\end{abstract}

\section{Introduction}%

\textit{Vision Transformers (ViTs)} \cite{Dosovitskiy2020,touvron2021training} have achieved state-of-the-art performance
in many computer vision tasks, yet they typically have high computational costs that scale quadratically with the number of tokens. However, in practice, a ViT's final prediction often depends on a subset of the most informative tokens~\cite{Chefer2021}, with the other tokens being redundant. This opens up the opportunity for token pruning to reduce the computation cost while maintaining high accuracy.

To date, several token pruning policies have been proposed, most of which~\cite{2023peeling,2022token,fayyaz2022adaptive,liange2022not,2022evo} draw on pre-defined, handcrafted pruning policies based on the attention values of the tokens, lacking adaptivity to varying inputs. Another existing work, DynamicViT~\cite{rao2021dynamicvit}, exploits Multi-Layer Perceptrons (MLPs) to automatically learn whether to prune each token. However, it still relies on manually defined pruning ratios for each layer, resulting in limited flexibility.
Furthermore, existing works largely overlook the sequential nature of the tokens, where the decision to preserve or prune certain tokens in one layer can significantly impact the outcomes for the subsequent layers.

\begin{figure}[t]
\begin{center}
\includegraphics[scale=0.3]{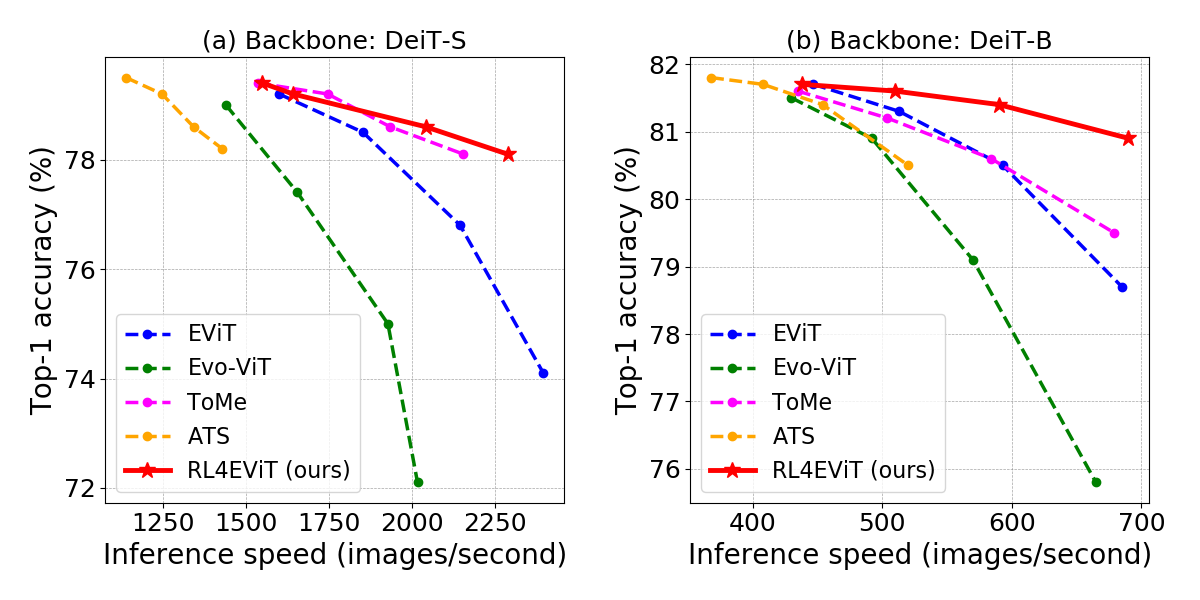}
\end{center}
\vspace{-15pt} 
\caption{Comparisons between our RL4EViT and some state-of-the-art token pruning methods on the ImageNet-1k dataset \cite{imagenet}, using DeiT-S and DeiT-B as backbones. Our RL4EViT has achieved the best trade-off between accuracy and inference speed.} 
\vspace{-15pt}
\label{fig:sota}
\end{figure}

\begin{figure*}[h]
\begin{center}
\includegraphics[width=.75\textwidth]{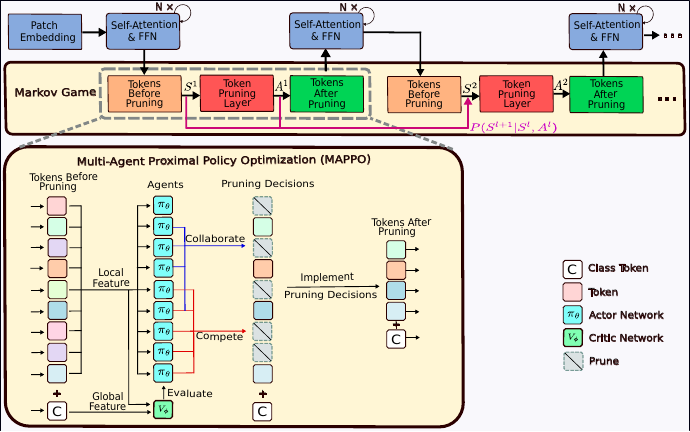}
\end{center}
\vspace{-10pt}
\caption{An illustration of our \textit{Reinforcement Learning for Efficient Vision Transformers (RL4EViT)} method for token pruning. For a given ViT, each Transformer block is proceeded by a token pruning layer, where we utilize Multi-Agent Proximal Policy Optimization (MAPPO) \cite{Yu2022} to prune the output tokens of the Transformer block. The global reward across all token decision layers is obtained via a Markov Game \cite{markovgame1994} to preserve important sequential information. Note that in practice, instead of directly removing the pruned tokens, we set their attention weights to -1000 to facilitate mini-batch-based training.
} 
\vspace{-15pt}
\label{fig:framework}
\end{figure*}

In view of the drawbacks of existing handcrafted features, in this work, we propose \textbf{Reinforcement Learning for Efficient Vision Transformer (RL4EViT)}, where we draw on \textit{Reinforcement Learning (RL)} to \textit{learn} a token pruning policy in a data-adaptive manner. To the best of our knowledge, this is the first work applying RL to this task, which can achieve the best trade-off between accuracy and efficiency when compared with the state-of-the-art (Fig. \ref{fig:sota}). Concretely, we formulate token pruning as a sequential decision problem. As is shown in Fig. \ref{fig:framework}, we incorporate a token pruning layer after each Transformer block\footnote{For simplicity but without loss of generality, in all sections in this paper but Section \ref{sec:expr}, we assume that all Transformer blocks are proceeded by token pruning layers. In practice, we can selectively add token pruning layers after a subset of Transformer blocks.}, where we utilize \textbf{Multi-Agent Proximal Policy Optimization (MAPPO)} \cite{Yu2022} to prune the output tokens of the Transformer block. The multi-agent setting allows for individualized treatment of each token as compared with a single-agent setting. We also propose reward functions that enable the agents to both compete and collaborate with each other, effectively balancing efficiency and accuracy. The global decision across all token pruning layers is obtained via a \textbf{Markov Game} \cite{markovgame1994}, thus preserving important inter-layer sequential information.


In summary, our main contributions are as follows:
\begin{itemize}[leftmargin=*]
    \item We propose RL4ViT where we exploit RL for data-adaptive token pruning in ViTs. To the best of our knowledge, this is the first time RL has been applied to this task.
    \item We present MAPPO-based token pruning layers, where the multi-agent setting enables individualized decision-making for each token. We propose novel reward functions for an effective trade-off between efficiency and accuracy, and utilize a Markov Game to calculate the global reward to capture sequential semantics across all token pruning layers.
    \item We conduct extensive experiments on the well-known ImageNet-1k dataset, which show that our RL4EViT can improve the inference speed by up to 44\% while incurring only a negligible accuracy drop of 0.4\%.
\end{itemize}
The rest of this paper is organized as follows: Section \ref{sec:review}
reviews the related work. Section
\ref{sec:method} details our RL4EViT method. Section \ref{sec:expr} reports the experimental results. Section \ref{sec:conc} concludes the paper.

\section{Related Work}
\label{sec:review}


Addressing their quadratic complexity against the number of tokens, researchers have proposed multiple token pruning methods for ViTs. Most existing works ~\cite{2023peeling, 2022token, fayyaz2022adaptive, liange2022not, 2022evo} favor preserving tokens with high attention weights over those with low ones. These methods all use pre-defined, handcrafted pruning policies that lack adaptivity to diverse inputs. DynamicViT~\cite{rao2021dynamicvit}, on the other hand, uses MLPs to predict whether to preserve or prune each token by balancing accuracy and efficiency. However, the number of tokens to prune in each layer remains manually defined, limiting its flexibility. 


Alongside token pruning, many methods entail \textit{token merging}~\cite{liange2022not,2022evo,2022token}. Instead of directly discarding less important tokens, these methods merge them into more compact representations using techniques such as spatial attention~\cite{liange2022not,2022evo} and pooling~\cite{2022token}. Such methods are more robust against errors in token pruning, at the cost of efficiency decay. 

\section{RL4EViT for Token Pruning}
\label{sec:method}
We are now in a position to present the details of our RL4EViT method, which integrates MAPPO token pruning layers into the ViT architecture and exploits a Markov Game setup to prune redundant tokens (Fig. \ref{fig:framework}). As illustrated in Alg.~\ref{alg:mappo4dynamicinference}, RL4EViT takes the following as the input:
\begin{itemize}[leftmargin=*]
    \item \textbf{image dataset $\mathcal{D}$};
    \item \textbf{ViT components}: patch embedding network $M$, Transformer blocks $F^l$ $(l = 1, 2, \ldots, L)$ entailing self-attention and Feed-Forward Networks (FFNs), and the image classifier $C$ (can be replaced with other downstream task solvers);
    \item \textbf{RL components}: an actor network $\pi_\theta$ and a critic network $V_\phi$, as core modules of MAPPO (detailed in Section \ref{sec:MAPPO}).
\end{itemize}
RL4ViT is trained based on a pre-trained ViT (line \ref{line:pretrain} of Alg.~\ref{alg:mappo4dynamicinference}) with no token pruning. Then, we alternate between optimizing the RL components and optionally fine-tuning the ViT components using the conventional forward-backward propagation scheme, eventually obtaining the token-pruned, fine-tuned ViT.
We now separately introduce the forward (lines \ref{line:patch_embed}-\ref{line:trajectory}) and backward (lines \ref{line:if_RL}-\ref{line:end_backward}) passes of RL4EViT.


\begin{algorithm}[t]
\caption{RL4EViT}\label{alg:mappo4dynamicinference}
\textbf{Input:} image dataset $\mathcal{D}$, patch embedding network $M$, Transformer blocks $F^l$ $(l = 1, 2, \ldots, L)$, classifier $C$, MAPPO actor network $\pi_\theta$, MAPPO critic network $V_\phi$ \\
\textbf{Output:} fine-tuned patch embedding network $M$, token-pruned Transformer blocks $F^l$ $(l = 1, 2, \ldots, L)$, fine-tuned classifier $C$
\begin{algorithmic}[1]
\State Pretrain $M$, $F^l$ $(l = 1, 2, \ldots, L)$, $C$ \label{line:pretrain}
\State Initialize $\pi_\theta$ and $V_\phi$ \label{line:initRL}

\For{each epoch $e = 1, 2, \dots, E$} \label{line:epoch}
    \For{each mini-batch $\mathcal{B} \in \mathcal{D}$} \label{line:minibatch}
        \LineComment{\textbf{Step 1: Forward Pass}}
    
        \State Obtain patch embedding $T^0 = M(\mathcal{B})$ \label{line:patch_embed}
        \For{each $F^l$ $l = 1, 2, \dots, L$} \label{line:mappo_for}
            \State Obtain tokens before pruning $S^l = F^l(T^{l-1})$ \label{line:unpruned}
            \State Obtain pruning decisions: $A^l$ = $\pi_\theta(S^l)$ \label{line:decide}
            \State Obtain tokens after pruning: $T^l$ = $S^l \times A^l$ \label{line:pruned}
            \State Update $F^l$ with $T^l$ \label{line:update_tokens}
        \EndFor \label{line:mappo_endFor}
        \State Compute rewards $R^1, R^2, \dots, R^l $ \label{line:rewards}
        \State Collect Markov Game trajectory $S^1$, $A^1$, $R^1$, $S^2$, $A^2$, $R^2$, $\dots$, $S^l$, $A^l$, $R^l$ \label{line:trajectory}
    
        \LineComment{\textbf{Step 2: Backward Pass}}
        \If{$e \% 2 == 1$} \Comment{Optimize RL with ViT fixed} \label{line:if_RL}
            \For{each RL iteration $k = 1, 2, \ldots, K$} \label{line:rl_iter}
                \State Compute policy objective $L_{\theta}$, value loss $\mathcal{L}_{\phi}$\label{line:rl_loss}
                \State Update $\pi_\theta$ and $V_\phi$ using $\mathcal{L}_{\theta}$ and $L_{\phi}$ \label{line:optimize_rl}
            \EndFor
        \Else \Comment{Fine-tune ViT with RL fixed} \label{line:if_vit}
            \State Compute classification loss $\mathcal{L_C}$ \label{line:main_loss}
            \State Update $M$, $F^l$ $(l = 1, 2, \ldots, L)$, $C$  with $\mathcal{L_C}$ \label{line:optimize_vit}
        \EndIf \label{line:end_backward}
    \EndFor \label{line:end_mini_batch}
\EndFor\\ \label{line:end_epoch}
\Return $M$, $F^l$ $(l = 1, 2, \ldots, L)$, $C$ \label{line:ret}
\end{algorithmic}
\end{algorithm}

\subsection{The Forward Pass}
\label{sec:forward}

In the forward pass (lines \ref{line:patch_embed}-\ref{line:trajectory} of Alg.~\ref{alg:mappo4dynamicinference}), we feed the tokens obtained from each Transformer block $F^l$ into a MAPPO token pruning layer, which takes pruning decisions $A^l$ using the MAPPO actor network $\pi_\theta$ for each token. The inputs, decisions and rewards of the multiple token pruning layers in the network can thus form a Markov Game trajectory, which will be used to optimize the RL components in the backward pass. We now separately present the MAPPO token pruning layers and the Markov Game trajectory.

\subsubsection{MAPPO Token Pruning Layers}
\label{sec:MAPPO}


The general idea of Multi-Agent Proximal Policy Optimization (MAPPO) \cite{Yu2022} for token pruning is to map each input token (except the class token, as in the context of image classification) to an agent. These agents, dictated by the actor network $\pi_\theta$, decide whether to prune their corresponding tokens. The pruning decisions are evaluated by the critic network $V_\phi$ from a global point of view. Compared with a single-agent setting where the pruning decisions of all tokens in the ViT are taken by a single agent, our multi-agent approach allows for more individualized and thus effective decision-making for each token. 

Concretely, we use MLPs for both the actor and critic networks. Furthermore, to reduce the number of learnable parameters, all agents in RL4EViT share a single actor network $\pi_\theta$ with identical parameters, and all token pruning layers share a single critic network $V_\phi$. Each agent takes in the feature vector of its corresponding token (which we call a \textit{local feature}) as the input, and outputs a binary decision value with 0 denoting pruning the token, and 1 denoting preserving it. The critic outputs a Q-value for the decision of each agent. Notably, while the MAPPO critic individually evaluates each agent, it does so from a \textit{global} point of view. Hence, it is vital to input into the critic an Agent-Specific Global Feature, fusing the agent-specific feature with an Environment-Provided \textit{global feature} \cite{Yu2022}. In RL4EViT, we use the downstream task-specific token (class token in image classification) as the global feature, concatenating it with the local feature as the input of the critic.

As with the reward function, we note that the goal of token pruning is to achieve the best trade-off between efficiency and accuracy. Accordingly, we define two reward functions for each agent. Specifically, given a single training image, for the $i$-th agent in layer $l$ $a^l_i$, the rewards are defined as 
\begin{equation}
\label{eq:rewards}
    \begin{aligned}
        r^l_{1i} & = 
        \begin{cases}
            0, & \text{if agent } a^l_i\text{ decides to \textit{prune} the token}, \\
            -1, & \text{if agent } a^l_i\text{ decides to \textit{preserve} the token}
        \end{cases}\\
        r^l_{2i} & = \frac{\gamma}{n^l}
    \end{aligned}
\end{equation}
where $n^l$ is the number of \textit{preserved} tokens (\textit{after} token pruning) in the $l$-th layer, and $\gamma$ is some accuracy metric for the downstream task. For image classification, we let $\gamma$ be 1 if the image is classified correctly, and 0 otherwise. The total reward for agent $a^l_i$ is
\begin{equation}
\label{eq:total_rewards}
R^l_i = \alpha*r^l_{1i} + \beta*r^l_{2i}
\end{equation}
where $\alpha$ and $\beta$ are pre-set weights.

Through the two rewards $r^l_{1i}$ and $r^l_{2i}$, we allow all agents (not only in a single token pruning layer, but across all layers) to both compete and collaborate to strike a balance between efficiency and accuracy. On the one hand, $r^l_{1i}$ incentivizes pruning as many tokens as possible, letting them compete for valuable computational cost budgets. On the other hand, $r^l_{2i}$ incentivizes them to collaborate to enhance the eventual downstream task accuracy. Notably, in $r^l_{2i}$, we amortize the accuracy gain to each preserved token by dividing $n^l$ to minimize the number of preserved tokens, keeping only the ones with the greatest contribution to the downstream task. 

Before we move on, we note that for a pruned token, we do not explicitly remove it from the ViT architecture,
as this would cause discrepancies in the network architecture from image to image, hindering mini-batch-based training.
Rather, following the practice of \cite{rao2021dynamicvit}, we mask the pruned tokens by setting their corresponding attention matrix values to a negative number with a large absolute value (e.g. -1000) before applying Softmax to them, effectively severing the connections between pruned and preserved tokens.

\subsubsection{The Markov Game Trajectory}
\label{sec:markov}
Token pruning for ViT is inherently sequential, as the tokens are obtained from consecutive Transformer blocks. Addressing this, we coordinate the multiple MAPPO token pruning layers using a Markov Game framework. Specifically, for the $l$-th ($l = 1, 2, \ldots, L$) layer, we are concerned with the following:
\begin{itemize}[leftmargin=*]
\item \textbf{state $S^l$:} feature vectors of the tokens before pruning;
\item \textbf{joint action $A^l$:} decisions by all agents in the $l$-th layer;
\item \textbf{reward $R^l$:} rewards (Eq.~\ref{eq:total_rewards}) of all agents in the $l$-th layer;
\end{itemize}
Assuming that the state of a layer $S^{l+1}$ depends solely on the state $S^l$ and the joint action $A^l$ of the previous layer, the \textbf{transition} from $S^l$ to $S^{l+1}$ is dictated by the conditional probability $P(S^{l+1}|S^l, A^l)$. The \textbf{trajectory} of the Markov Game is defined as
$$
   S^1, A^1, R^1, S^2, A^2, R^2, ..., S^L, A^L, R^L
$$
which will be used to calculate the RL loss functions.









\subsection{The Backward Pass}
\label{sec:back}
In the backward pass (lines \ref{line:if_RL}-\ref{line:end_backward} of Alg.~\ref{alg:mappo4dynamicinference}), we alternately optimize the RL components and fine-tune the ViT.

\subsubsection{Optimizing the RL components}
\label{sec:optimize_rl}
The RL components to optimize are the actor network $\pi_\theta$ and the critic network $V_\phi$. Specifically, we can obtain a Markov Game trajectory for each training image, resulting in $B = |\mathcal{B}|$ trajectories for the current mini-batch $\mathcal{B}$. We iteratively optimize $\pi_\theta$ and $V_\phi$ based on these trajectories. In each iteration, for $\pi_\theta$, we maximize the following MAPPO policy objective \cite{Yu2022}
\begin{equation}
\label{eq:l_theta}
L_\theta = \frac{1}{\sum_{b=1}^{B} M^b}
\sum_{b=1}^{B}\sum_{m=1}^{M^b}  \min\big( 
    \mathtt{r}^b_{\theta, m} \mathcal{\hat{A}}^b_m, \text{clip}(\mathtt{r}^b_{\theta, m}, 1\pm\epsilon_\theta) \mathcal{\hat{A}}^b_m 
\big)
\end{equation}
where for the $m$-th agent among the $M^b$ agents involved in the $b$-th trajectory, $\mathcal{\hat{A}}^b_m$ is its advantage obtained via Generalized Advantage Estimation (GAE)~\cite{Schulman2015a}.
$\mathtt{r}^b_{\theta, m}
=\pi_{\theta}^{\text{new}}(a^b_m \mid
s_m^b) / \pi_\theta^{\text {old }}(a^b_m \mid
s_m^b)$ where $s_m^b$ is the state of the agent (i.e. the feature vector of its input token), $\pi_{\theta}^{\text{new}}$ and $\pi_{\theta}^{\text{old}}$ are the current $\pi_{\theta}$ and that at the end of the previous epoch, $a^b_m = \pi_{\theta}^{\text{new}}(s_m^b)$ is the action (i.e. \textit{prune} or \textit{preserve}) to take according to the current $\pi_{\theta}$, and $\pi_{\theta}(a^b_m \mid s_m^b)$ is the logit (probability) that corresponds to $a^b_m$ outputted by $\pi_{\theta}$ (new or old). $\epsilon_\theta$ is a bounding hyperparameter for $\mathtt{r}^b_{\theta, m}$.

For the critic network $V_\phi$, we minimize the following MAPPO value loss \cite{Yu2022}:
\begin{align}
\label{eq:l_phi}
L_\phi = &\frac{1}{\sum_{b=1}^{B} M^b} \sum_{b=1}^{B} \sum_{m=1}^{M^b} \max\Big[ 
    \big(V^{\text{new}}_\phi(\mathtt{s}^{b}_m) - R^b_m\big)^2, \nonumber \\
    &\big(\text{clip}(V^{\text{old}}_\phi(\mathtt{s}^{b}_m), V^{\text{old}}_\phi(\mathtt{s}^{b}_m) \pm \epsilon_\phi) - R^b_m\big)^2
\Big]
\end{align}
where for the $m$-th agent, $\mathtt{s}^{b}_m$ is its global feature (i.e. the concatenation of the feature vectors of its input token and the class token), $\mathtt{R}^{b}_m$ is its reward, with $V^{\text{new}}_\phi$ and $V^{\text{old}}_\phi$ being the current $V_\phi$ and that at the end of the previous epoch. $\epsilon_\theta$ is a bounding hyperparameter.

\subsubsection{Fine-tuning the ViT}
After token pruning, the pre-trained ViT parameters may no longer be optimal. Therefore, we allow for optionally fine-tune \textit{all} the ViT parameters after each update to the token pruning policy. The effectiveness of fine-tuning will be empirically shown in Section \ref{sec:expr_ablation}.

\section{Experiments}
\label{sec:expr}
We now report our experimental results, obtained on the well-known ImageNet-1k~\cite{imagenet} dataset. Following the practices of previous works~\cite{rao2021dynamicvit,2023peeling,liange2022not,2022evo,2022token,fayyaz2022adaptive}, we instantiate RL4EViT for the image classification task using DeiT-S and DeiT-B~\cite{touvron2021training} as the backbone ViTs. Both backbones have 12 Transformer blocks. Following \cite{rao2021dynamicvit}, we only conduct token pruning for blocks 3, 6 and 9 in RL4EViT. The actor network $\pi_\theta$ is an MLP with five layers of the sizes (384, 1536, 256, 64, 2) for DeiT-S and (768, 3072, 256, 64, 2) for DeiT-B. The critic network $V_\phi$ is an MLP with five layers of the sizes (768, 1536, 256, 64, 1) for DeiT-S and (1536, 3072, 256, 64, 1) for DeiT-B. The number of RL training iteration $K$ in line~\ref{line:rl_iter} of Alg.~\ref{alg:mappo4dynamicinference} is 15. Both $\pi_\theta$ and $V_\phi$ are optimized with Adam with a learning rate of $5 \times 10^{-5}$. The other hyperparameters are the same as those in \cite{Yu2022}. All experiments are run on an NVIDIA A6000 GPU.



\subsection{Comparison Against Rival Methods}
\label{expr:rivals}
\begin{figure}[t]
\begin{center}
\includegraphics[scale=0.2]{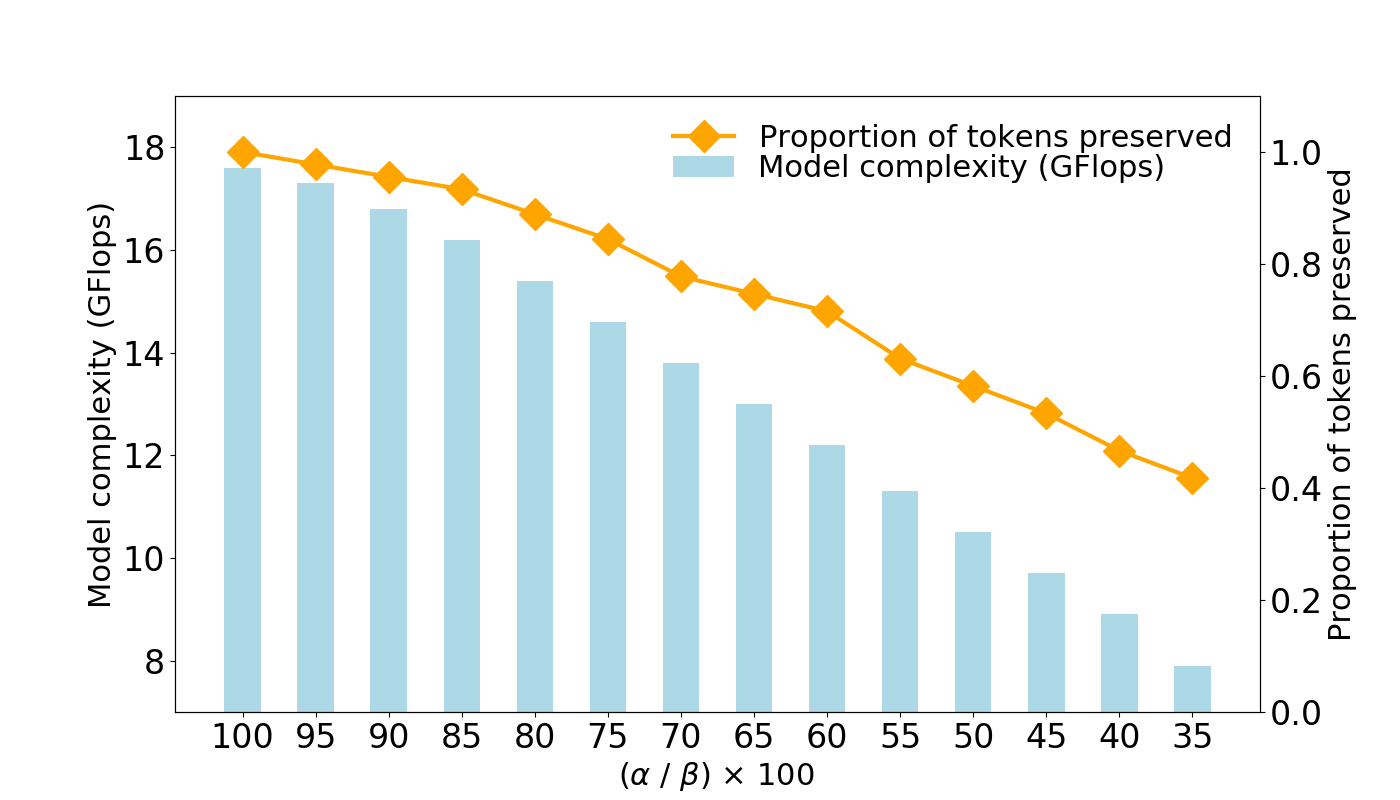}
\end{center}
\vspace{-15pt}
\caption{Relationship between the proportion of the preserved tokens, as well as the model complexity, and $\alpha / \beta$, with DeiT-B as the backbone ViT.}

\vspace{-15pt}
\label{fig:alpha_beta}
\end{figure}

We compare the accuracy (measured by \textit{top-1 accuracy}) and efficiency (measured by \textit{inference speed}) of our RL4EViT against multiple state-of-the-art methods~\cite{rao2021dynamicvit,2023peeling,liange2022not,2022evo,2022token,fayyaz2022adaptive}. Following the practices of previous works~\cite{rao2021dynamicvit}, we make sure that all methods compared have approximately the same model complexity (quantified by GFLOPs) for fair comparison.
For our RL4EViT, the model complexity can be empirically adjusted by tuning $\alpha$ and $\beta$ in Eq.~\ref{eq:total_rewards}. Specifically, the larger $\alpha / \beta$ is, the larger the number of preserved tokens, and the larger the GFLOPs. While there is no way to mathematically formulate the exact relationship between the GFLOPs and $\alpha / \beta$, this can be empirically dipicted by Fig.~\ref{fig:alpha_beta}. We can thus vary $\alpha / \beta$ to obtain models with different GFLOPs. 
Also, note that while our method allows for fine-tuning of the ViT, not all the rival methods allow for this. To be fair to these methods, we disregard fine-tuning in this experiment.

\begin{table*}[htbp]
\centering
\caption{Comparison of RL4EViT against rival methods. Top-1 accuracies are formatted as "\textit{Absolute value (decay from no token pruning)}", best results are in \textbf{bold}, second bests are \underline{underlined}. Inference speeds are formatted as "\textit{Absoulte value (speedup against no token pruning)}".}
\vspace{-8pt}
\label{tab:deits}
\resizebox{\textwidth}{!}{%
\begin{tabular}{@{}lccccccc@{}}
\toprule
\multicolumn{8}{c}{\textbf{Backbone: DeiT-S} (Top-1 Acc. 79.9\% at 1265.3 images/s with \textit{no token pruning})}\\
\midrule
\multirow{2}{*}{\textbf{Method}} & \multirow{2}{*}{\textbf{Params}}  &
\multicolumn{2}{c}{\textbf{Approx. 3.5 GFLOPs $(\alpha/\beta=0.80)$}} & \multicolumn{2}{c}{\textbf{Approx. 3.0 GFLOPs $(\alpha/\beta=0.72)$}} & \multicolumn{2}{c}{\textbf{Approx. 2.6 GFLOPs $(\alpha/\beta=0.56)$}} \\ 
\cmidrule(lr){3-4} \cmidrule(lr){5-6} \cmidrule(lr){7-8}
                        &                     &  \textbf{Top-1 Acc. (\%)}    & \textbf{Speed (images/s)}    &  \textbf{Top-1 Acc. (\%)}    & \textbf{Speed (images/s)}     &  \textbf{Top-1 Acc. (\%)}    & \textbf{Speed (images/s)} \\ \midrule
DynamicViT~\cite{rao2021dynamicvit}             & 22.8M                  & 74.0 (-5.9)          & 1478.4 (×1.17)   & 67.4 (-12.5)             & 1700.1 (×1.34)     & 58.3 (-21.6)             & 1980.9 (×1.57) \\ 
Tri-Level~\cite{2023peeling}               & 22.1M               & 67.6 (-12.3)             & 1345.9 (×1.06)     & 67.6 (-12.3)            & 1551.2 (×1.23)    & 67.6 (-12.3)            & 1793.8 (×1.41) \\ 
EViT~\cite{liange2022not}             & 22.1M       & 79.2 (-0.7)             & 1600.5 (×1.26)   & 78.5 (-1.4)             & 1852.3 (×1.46)     & 76.8 (-3.1)             & 2144.5 (×1.69) \\ 
Evo-ViT~\cite{2022evo}            & 22.1M            & 79.0 (-0.9)           & 1439.3 (×1.14)    & 77.4 (-2.5)            & 1655.4 (×1.31)    & 75.0 (-4.9)             & 1927.9 (×1.52)\\ 
ToMe~\cite{2022token}              & 22.1M                & \underline{79.4 (-0.5)}           & 1536.7 (×1.21)    & \textbf{79.2 (-0.7)}              & 1746.0 (×1.38)     & \textbf{78.6 (-1.3)}              & 1934.0 (×1.53)  \\ 
ATS~\cite{fayyaz2022adaptive}               & 22.1M            & \textbf{79.5 (-0.4)}             & 1140.6 (×0.90)    & \textbf{79.2 (-0.7)}     & 1248.1 (×0.99)   & \textbf{78.6 (-1.3)}     & 1343.0 (×1.06) \\ 
\vspace{2pt}
RL4EViT (ours)               & 22.3M                 & \underline{79.4 (-0.5)}              & 1550.4 (×1.23)   & \textbf{79.2 (-0.7)}     & 1641.8 (×1.30)   & \textbf{78.6 (-1.3)}     & 2041.4 (×1.61) \\ 

\toprule
\multicolumn{8}{c}{\textbf{Backbone: DeiT-B} (Top-1 Acc. 81.8\% at 408.8 images/s with \textit{no token pruning})}\\
\midrule
\multirow{2}{*}{\textbf{Method}} & \multirow{2}{*}{\textbf{Params}} & 
\multicolumn{2}{c}{\textbf{Approx. 15.3 GFLOPs $(\alpha/\beta=0.88)$}} & \multicolumn{2}{c}{\textbf{Approx. 13.1 GFLOPs $(\alpha/\beta=0.76)$}} & \multicolumn{2}{c}{\textbf{Approx. 11.6 GFLOPs $(\alpha/\beta=0.68)$ }} \\ 
\cmidrule(lr){3-4} \cmidrule(lr){5-6} \cmidrule(lr){7-8}
                        &                        &  \textbf{Top-1 Acc. (\%)}    & \textbf{Speed (images/s)}     &  \textbf{Top-1 Acc. (\%)}    & \textbf{Speed (images/s)}     &  \textbf{Top-1 Acc. (\%)}    & \textbf{Speed (images/s)} \\ \midrule
DynamicViT~\cite{rao2021dynamicvit}             & 89.5M                    & 79.9 (-1.9)             & 420.0 (×1.03)    & 77.7 (-4.1)             & 489.0 (×1.20)    & 75.5 (-6.3)             & 579.7 (×1.42)  \\ 
Tri-Level~\cite{2023peeling}           & 86.6M                      & 64.6 (-17.2)             & 398.9 (×0.98)     & 64.6 (-17.2)             & 466.7 (×1.14)    & 64.6 (-17.2)            & 539.7 (×1.32) \\ 
EViT~\cite{liange2022not}             & 86.6M                   & \underline{81.7(-0.1)}              & 447.2 (×1.09)   & 81.3 (-0.5)             & 513.0 (×1.25)    & 80.5 (-1.3)             & 593.2 (×1.45)  \\ 
Evo-ViT~\cite{2022evo}          & 86.6M                   & 81.5 (-0.3)             & 430.6 (×1.05)   & 80.9 (-0.9)            & 492.2 (×1.20)   & 79.1 (-1.7)            & 570.2 (×1.39) \\ 
ToMe~\cite{2022token}            & 86.6M                      & 81.6 (-0.2)            & 435.1 (×1.06)    & 81.2 (-0.6)              & 504.9 (×1.24)    & 80.6 (-1.2)             & 584.5 (×1.43) \\ 
ATS~\cite{fayyaz2022adaptive}               & 86.6M                      & \textbf{81.8 (-0.0)}              & 368.3 (×0.90)   & \textbf{81.7 (-0.1)}    & 407.9 (×1.00)    & \textbf{81.4 (-0.4)}    & 453.7 (×1.10) \\ 
RL4EViT (ours)                & 86.8M                    & \underline{81.7 (-0.1)}             & 438.3 (×1.08)    & \underline{81.6 (-0.2)}     & 509.9 (×1.25)    & \textbf{81.4 (-0.4)}     & 589.8 (×1.44) \\
\bottomrule
\vspace{-15pt}
\end{tabular}}
\end{table*}


The results are shown in Table \ref{tab:deits}, which indicates that our RL4EViT can generally achieve the best trade-off between efficiency and accuracy. Specifically, most rival methods~\cite{2023peeling,liange2022not,2022evo,2022token,fayyaz2022adaptive} apply purely handcrafted pruning policies, lacking adaptivity to diver input data. The rival method that is most affine to ours is DynamicViT~\cite{rao2021dynamicvit}, which exploits MLPs to learn the pruning policy. However, its flexibility is still limited by the fact that the number of tokens to prune in each layer remains manually defined, limiting its performance. By contrast, by fully unleashing the power of learned token pruning policies with RL, our RL4EViT can improve the inference speed by up to 44\% while incurring a neglectable accuracy decay of only 0.4\% as compared with no token pruning (in the case where the backbone is DeiT-B and the approximate GFLOPs is 11.6), which can be further improved to just 0.1\% of accuracy decay when we apply fine-tuning to the ViT.
Also note that several rival methods~\cite{liange2022not,2022evo,2022token} entail separate token merging modules (see Section~\ref{sec:review}) to enhance the accuracy. By contrast, our RL4EViT can outperform or match their accuracy in all cases with no need for token merging.





\subsection{Ablation Studies}
\label{sec:expr_ablation}
We now conduct ablation studies for certain design features of our method, of which the most important one is our multi-agent setting using MAPPO within the framework of a Markov Game. To validate its effectiveness, we consider the following two alternative token pruning policies: a \textbf{Random} policy where tokens are randomly pruned, and a \textbf{single-agent PPO} \cite{ppo} policy which, instead of using an agent for each token, uses a single agent to make the pruning decisions for all tokens in the ViT. The results are shown in Table \ref{tab:whymappo}. Unsurprisingly, the random policy is significantly inferior to the RL-based policies. Moreover, our multi-agent method outperforms its single-agent counterpart in all cases, owing to the former's ability to make a more individualized decision for each token.


\begin{table}[t]
    \caption{Comparison of top-1 accuracies with different token pruning policies, which validates the effectiveness of our multi-agent method with MAPPO.}
    \vspace{-12pt}
    \label{tab:whymappo}
    \begin{center}
        \begin{tabular}{lcccc}
            \toprule
             \multirow{2}{*}{\textbf{Backbone}}  & \multirow{2}{*}{\textbf{GFLOPs}}& \multicolumn{3}{c}{\textbf{Top-1 Acc. (\%) of Different Policies}}  \\ \cmidrule(lr){3-5} & &\textbf{Random} & \textbf{PPO} & \makecell{\textbf{MAPPO}\\\textbf{(ours)}} \\
            \midrule
            \multirow{2}{*}{DeiT-S}
                                     & 2.1       &35.8         &  76.5  &\textbf{76.8} \\
                                   \cmidrule{2-5}
                                     & 2.6   & 47.1               & 78.3  &\textbf{78.6} \\
                                
            \midrule
            \multirow{2}{*}{DeiT-B}           
                                   
                                    & 10.8   &  50.4      &  80.9 &\textbf{81.3} \\
                           
                                   \cmidrule{2-5}
                                    & 11.6    &  61.5            &  81.0  &\textbf{81.4} \\

            \bottomrule
        \end{tabular}
    \end{center}
    \vspace{-15pt}
\end{table}

\begin{figure}[t]
\begin{center}
\includegraphics[scale=0.19]{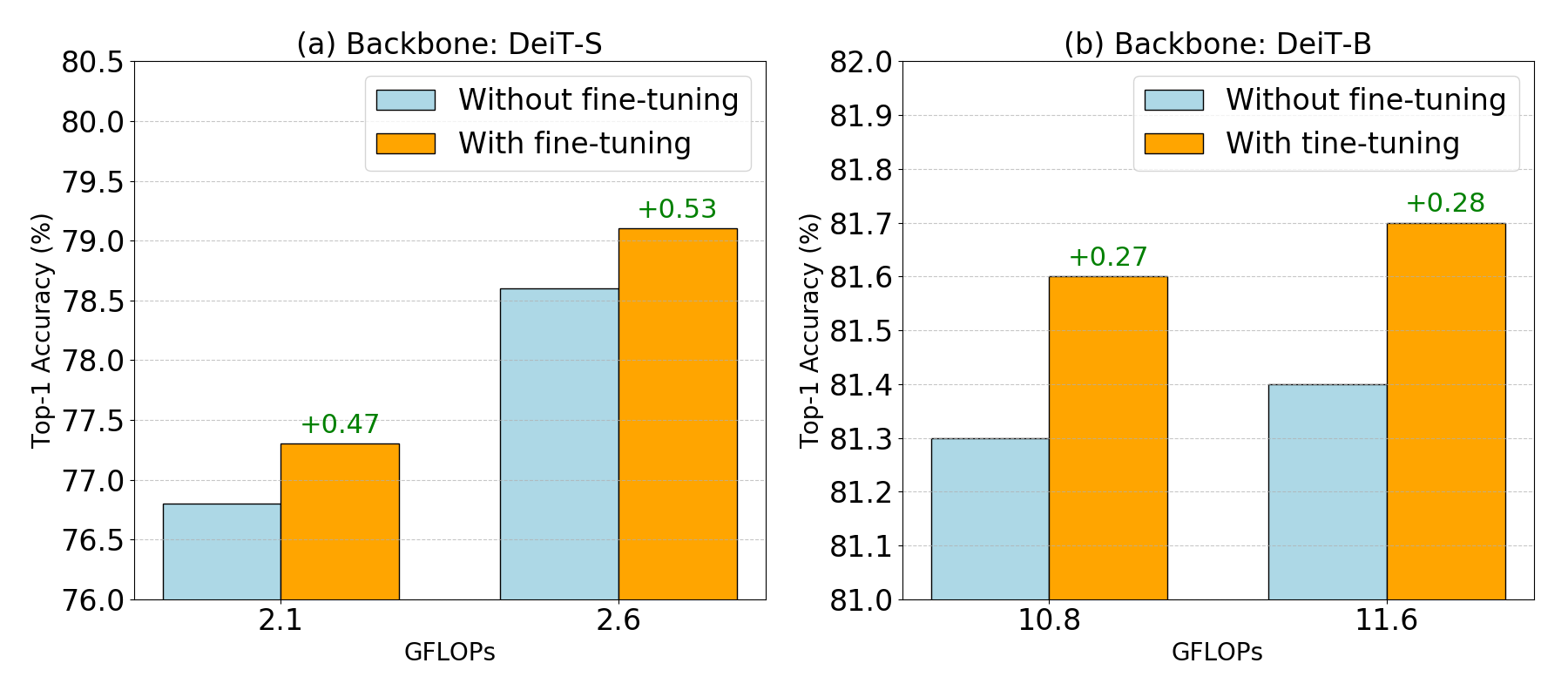}
\end{center}
\vspace{-15pt}
\caption{Top-1 accuracies with and without ViT fine-tuning}

\vspace{-15pt}
\label{fig:finetune}
\end{figure}

Another design choice of ours is optional fine-tuning for the ViT after each update to the token pruning policy. Fig.~\ref{fig:finetune} shows the accuracies with and without fine-tuning, indicating that fine-tuning can make the ViT more adapted to the updated network configuration, yielding better accuracies.


\subsection{Visualization of Token Pruning Results}

\begin{figure*}[h]
    \begin{center}
        \includegraphics[scale=0.87]{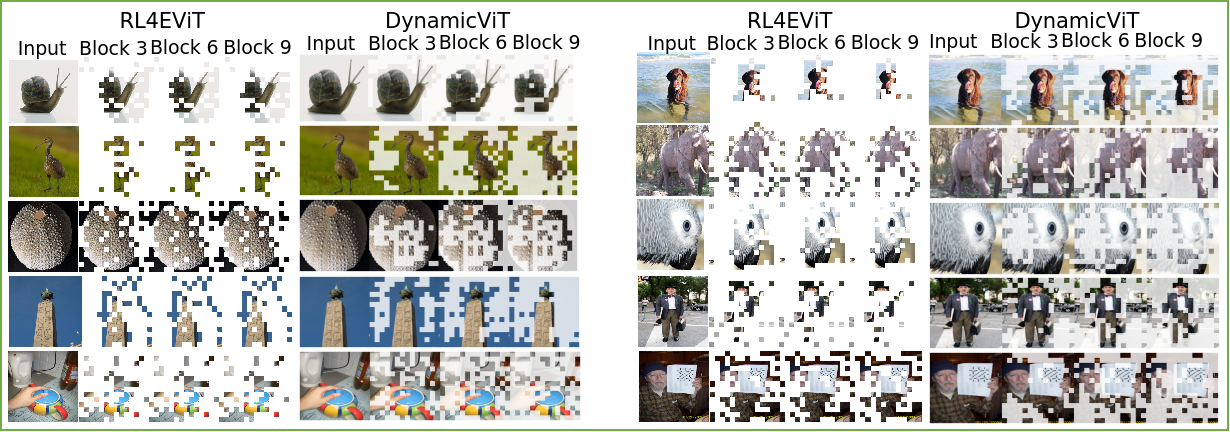}
    \end{center}
    \vspace{-10pt}
    \caption{Visualization of token pruning results obtained via our RL4EViT and DynamicVit~\cite{rao2021dynamicvit}, after Transformer blocks 3, 6, and 9 in DeiT-B~\cite{touvron2021training}. The pruned tokens are shown in white. As is indicated, thanks to its more data-adaptive nature, our RL4EViT can conduct most of the pruning at the earliest possible stage (block 3), leaving only the most important tokens behind. By contrast, DynamicViT can only prune a fixed number of tokens in each block, which limits its flexibility and thus its performance.}
    \vspace{-15pt}
\label{fig:visual}
\end{figure*}

To further demonstrate the effect of the token pruning policy in our RL4EViT, we conduct a visual comparison on several images in the ImageNet-1k dataset between it and DynamicVit~\cite{rao2021dynamicvit}, which as mentioned previously is the rival method most similar to ours due to its learned (rather than purely handcrafted) nature. Specifically, in Figure \ref{fig:visual}, we show the original input image and the token pruning results after Transfomer blocks 3, 6, and 9 in DeiT-B~\cite{touvron2021training}, where
the pixels corresponding to the pruned tokens are shown in white.

Note that our RL4EViT prunes most of the tokens in the first stage (block 3), while DynamicViT conducts the pruning in a more gradual manner. This is because in DynamicViT, the number of tokens pruned in each layer is fixed to a manually defined value, while our RL4EViT can automatically learn the number of tokens to prune in each layer in an attempt to strike the best global balance between accuracy and efficiency. As a result, it tends to prune unimportant tokens as early as possible, only passing the most important ones to later layers. Moreover, in most cases, the quality of the tokens preserved by our method is better than that by DynamicViT. For instance, for the third example in the second column of Fig. \ref{fig:visual} which is an image of a bird's head, our RL4EViT can preserve the tokens corresponding to the eye of the bird, while DynamicViT fails to do so, showcasing the superiority of our method.

\section{Conclusion}
\label{sec:conc}
In this paper, we proposed RL4EViT, which is (to the best of our knowledge) the first RL-based token pruning method for ViTs. By exploiting multiple MAPPO token pruning layers under a Markov Game framework, our method can be highly adaptive to varying input data, and capture important sequential information. Extensive experiments have shown that it can achieve state-of-the-art performance in terms of the trade-off between efficiency and accuracy. For future work, we will further study the application of our method in diverse downstream tasks other than image classification.

\section*{Acknowledgment}
\label{sec:acknowledgment}
Work funded by the Science Research Project of Hebei Education Department under Grant HJYB202516, the Beijing-Tianjin-Hebei Fundamental Research Cooperation Project under Grant F2024210051, the Data Intelligence Institute of Paris (diiP), and IdEx Université Paris Cité (ANR-18-IDEX-0001).



\bibliographystyle{IEEEbib}
\bibliography{rl4evit}
\end{document}